\documentclass[10pt,twocolumn,letterpaper]{article}

\usepackage{iccv}              

\definecolor{iccvblue}{rgb}{0.21,0.49,0.74}
\usepackage[pagebackref,breaklinks,colorlinks,allcolors=iccvblue]{hyperref}
\usepackage{balance}

\def\paperID{3171} 
\def\confName{ICCV}
\def\confYear{2025}

\def\projname{Group Inertial Poser\xspace}
\def\projshort{GIP\xspace}
\def\datasetname{GIP-DB\xspace}

\title{Group Inertial Poser: Multi-Person Pose and Global Translation \\ from Sparse Inertial Sensors and Ultra-Wideband Ranging}

\author{Ying Xue, Jiaxi Jiang, Rayan Armani, Dominik Hollidt, Yi-Chi Liao, and Christian Holz\\
Department of Computer Science, ETH Zürich, Switzerland\\[.2em]
{\small\href{https://siplab.org/projects/GroupInertialPoser}{\color{magenta}{\texttt{https://siplab.org/projects/GroupInertialPoser}}}}%
\vspace{-4mm}%
}

\begin{document}

\twocolumn[{%
\renewcommand\twocolumn[1][]{#1}%
\maketitle
\begin{center}
    \newcommand{\teaserwidth}{\textwidth}
    \centerline{
    \includegraphics[width=\linewidth]
    {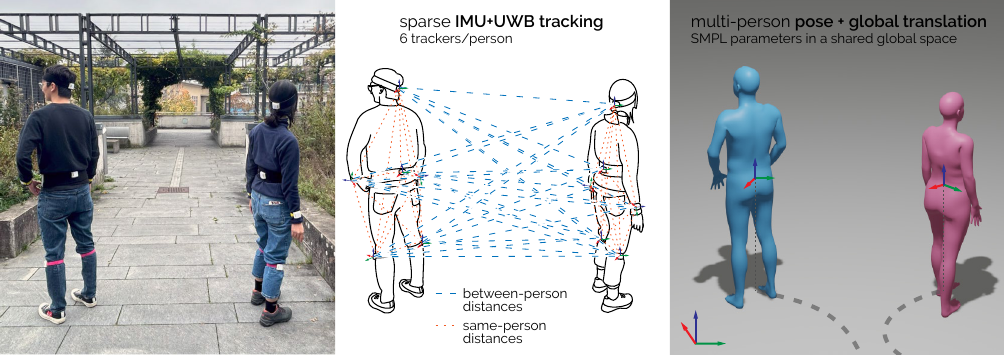}}
  \vspace{-1mm}%
  \captionof{figure}{
  Our method \emph{Group Inertial Poser} estimates 3D full-body poses and global translation for multiple humans using inertial measurements from a sparse set of wearable sensors, augmented by the distances between the sensors via ultra-wideband ranging.
  Our approach overcomes the challenge of drift in previous inertial pose estimators to track translation, as we leverage information from body motions \emph{across} multiple people.
  Our IMU+UWB method stabilizes and improves individual pose estimates, relative translation estimates between people, and global translation estimates, thereby preserving meaningful interaction dynamics.
  }%
\label{fig:teaser}%
\vspace{-1mm}%
\end{center}%
}]

\maketitle

\begin{abstract}
Tracking human full-body motion using sparse wearable inertial measurement units (IMUs) overcomes the limitations of occlusion and instrumentation of the environment inherent in vision-based approaches.
However, purely IMU-based tracking compromises translation estimates and accurate relative positioning between individuals, as inertial cues are inherently self-referential and provide no direct spatial reference for others.
In this paper, we present a novel approach for robustly estimating body poses and global translation for multiple individuals by leveraging the distances between sparse wearable sensors---both on each individual and across multiple individuals.
Our method \emph{Group Inertial Poser} estimates these absolute distances between pairs of sensors from ultra-wideband ranging (UWB) and fuses them with inertial observations as input into structured state-space models to integrate temporal motion patterns for precise 3D pose estimation.
Our novel two-step optimization further leverages the estimated distances for accurately tracking people's global trajectories through the world. 
We also introduce \emph{GIP-DB}, the first IMU+UWB dataset for two-person tracking, which comprises 200~minutes of motion recordings from 14~participants. 
In our evaluation, Group Inertial Poser outperforms previous state-of-the-art methods in accuracy and robustness across synthetic and real-world data, showing the promise of IMU+UWB-based multi-human motion capture in the wild. 
Code, models,  dataset: \href{https://github.com/eth-siplab/GroupInertialPoser}{github.com/eth-siplab/GroupInertialPoser}.

\end{abstract}

\section{Introduction}
\label{sec:intro}

Accurate motion tracking is a long-standing goal in computer vision. 
Single-person motion tracking has been extensively studied using camera-based approaches~\cite{kanazawa2018end, chen20173d, kocabas2020vibe, habermann2019livecap, shin2024wham, wang2024tram, bogo2016keep}.
Extending such tracking to multi-person motions introduces computational complexity and new challenges.
While vision-based approaches are accurate~\cite{optitrack, vicon, alphapose, li2020simple, kreiss2019pifpaf, cao2017realtime, wang2025prompthmr}, they struggle with occlusion in crowded environments and face challenges when individuals are close.
Their frequent reliance on stationary camera setups further constrains the effective tracking range.

Body-worn sensor-based approaches offer a promising alternative to these challenges.
Using sparse sets of wearable motion sensors, typically inertial measurement units (IMUs), has become popular to capture individuals' movements independent of environmental factors such as lighting, occlusion, or dynamic backgrounds.
Recent IMU-based approaches~\cite{yi2022physical,yi2021transpose,yi2024pnp} estimate body poses from six body-worn sensors in controlled environments, but they often exhibit drift in their predictions and, thus, struggle with estimating global translation.
Extending these approaches to multi-person constellations would further complicate accurately estimating relative positions \emph{between} individuals.
This prevents them from capturing inter-personal dynamics and spatial relationships between people---aspects that are highly interesting to reconstruct when aiming to understand interactions in real social scenarios.


In this paper, we introduce \emph{\projname} (GIP), a novel approach for robust multi-person 3D pose and global translation estimation from sparse inertial sensing with inter-sensor distances (as illustrated in Figure~\ref{fig:teaser}).
It first uses structured state-space models to estimate individual body poses and translations from inertial signals and same-person sensor distances. Then, a two-step optimization process refines the translation estimates:
(a) Relative position optimization aligns individuals in a shared world frame using between-person distances, eliminating the need for calibrated starting positions;
(b) Trajectory optimization further improves global translation accuracy.

To validate \projshort in \emph{real-world settings}, we introduce \emph{\datasetname}, a novel motion dataset that captures diverse activities from pairs of 14 participants who interacted during recording. 
Evaluating \projshort on \datasetname, our results show that \projshort estimates more accurate poses and translations with reduced drift---despite high noise levels in between-person distances.
Finally, we demonstrate \projshort for a four-person setting, where our method's performance improves as it leverages more distances.
By analyzing the reconstructed inter-human motions, we demonstrate that \projshort effectively captures relative spatial relationships and preserves meaningful inter-personal motion dynamics---an essential capability for modeling human-to-human interaction that previous approaches fail to achieve. 

\subsection*{Contributions}

\begin{enumerate}[leftmargin=*,nosep]
    \item \emph{\projname} (\projshort), a novel method to incorporate between-sensor distances and inertial signals to estimate 3D full-body poses and global translations for multiple people.
    \emph{\projshort} is the first IMU+UWB-based solution for estimating multi-person motion in a \emph{shared reference frame} and sets new state-of-the-art results.

    \item A structured state-space network that efficiently models sequential data to improve human pose estimation.
    To our knowledge, we are the first to adapt state-space models for inertial-based human motion estimation.

    \item An optimization-based method to estimate people's initial world positions, allowing tracked users to start at arbitrary locations and eliminating the need for calibration, manual setup, or aligning synchronized motions.
        
    \item \emph{\datasetname}, the first IMU+UWB two-person motion dataset with diverse activities from 14 participants who interact and perform everyday movements, totaling over 200~minutes of captured motions.
    \datasetname comprises synchronized IMU signals, UWB-based distance measurements, and SMPL body motion parameters.


\end{enumerate}
\section{Related Work}
\label{sec:related_work}

\paragraph{Multi-Person Motion Capture.}
Human motion capture, for both single and multiple individuals, has traditionally relied on camera-based approaches, with marker-based systems~\cite{optitrack, vicon} offering high precision but requiring costly setups confined to indoor environments. 
Recent advances in computer vision have enabled human pose estimation from sparse images~\cite{kanazawa2018end, bogo2016keep, chen20173d, multi-hmr2024} or videos~\cite{kocabas2020vibe, habermann2019livecap, ugrinovic2024multiphys, humanMotionKZFM19, wang2025prompthmr, zhang2024rohm}.
Frames from third-person~\cite{hu2021conditional}, floor~\cite{branzel2013gravityspace,augsten2010multitoe}, or egocentric cameras~\cite{jiang2024egoposer,tome2019egopose} extend naturally to multi-person scenarios, with challenges mainly in accurate person detection, tracking, and pose estimation.
Efforts such as RTMO~\cite{Lu2023RTMO} and HigherHRNet~\cite{Cheng2019HigherHRNetSR} proposed efficient bottom-up approaches for multi-actor tracking, while PETR~\cite{shi2022endtoend} uses transformer-based models and AlphaPose~\cite{alphapose, li2019crowdpose} offers joint pose estimation-tracking frameworks. 
However, tracking in crowded or occluded scenarios remains challenging due to obstruction and inconsistent tracking~\cite{Doring2022posetrack}, which wearable inertial motion capture systems can address by overcoming subject identification and occlusion issues.

\begin{figure*}[t!]
    \centering
    \includegraphics[width=\linewidth]
    {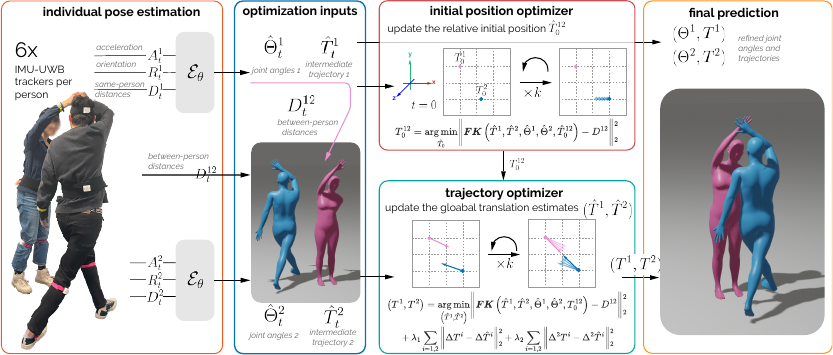}
    \caption{Overview of \projname (\projshort). Our pipeline consists of three key steps. It begins with individual pose estimation using an SSM-based model $\mathbf{\mathcal{E}_\theta}$ to generate a full-body SMPL pose and translations. The optimization steps then refine these estimates by minimizing the discrepancy between predicted and actual between-sensor distances. First, \projshort optimizes the initial relative positions ($T_0^{12}$); second, it fine-tunes the translations for both users ($T^1$, $T^2$).
}%
    \label{fig:pipeline}%
\vspace{-2mm}%
\end{figure*}

\paragraph{Motion Capture with Wearable Sensors.}

Wearable inertial sensors (IMU) have emerged as an alternative approach to motion capture, given their low power, small form factor, and affordability. Commercial systems, such as XSens~\cite{XSens} or Noitom~\cite{noitom} use 17 to 19~IMUs with underlying biomechanical models to estimate body pose.
With the availability of large motion capture datasets~\cite{AMASS:ICCV:2019, trumble2017total}, learning-based methods using only sparsely worn IMUs~\cite{von2017sparse, huang2018deep, nagaraj2020rnn, ilic2025human, mollyn2023imuposer} are emerging. Recent approaches use 6~IMUs and estimate both body pose and translation through ground contact points~\cite{yi2021transpose, jiang2022transformer} and physical constraints~\cite{yi2025improving, yi2022physical, yi2024pnp}. 
Several methods have further reduced input requirements to just the upper body alone~\cite{yang2021lobstr, ahuja2021coolmoves, jiang2022avatarposer, jiang2024egoposer, jiang2024manikin, starke2024categorical, ma2024nymeria, guzov2024hmd, li2023ego, wang2023scene}.

IMUs detect relative acceleration and angular velocity, so IMU-only methods have lower pose estimation and translation accuracy compared to vision-based alternatives. 
To address this, researchers have explored hybrid approaches incorporating external~\cite{pan2023fusingmono, von2018recovering, streli2023hoov} or body-worn~\cite{yi2023EgoLocate, hollidt2024egosim, li2020mobile} cameras. 
Alternative sensing modalities have also been investigated, including wearable ultrasonic~\cite{vlasic2007practical, liu2011realtime} and electromagnetic sensors~\cite{kaufmann2021pose, kaufmann2023emdb}. 
Recently, UIP~\cite{UIP} proposed leveraging distance measurements from ultra-wideband ranging to constrain IMU drift. 
Yet, these approaches focus on single-person tracking as sensors worn by different actors are treated as disjoint problems. 
We propose a novel approach to IMU+UWB-based pose estimation that, for the first time, enables multi-human relative pose estimation solely from inertial sensors and UWB. 

\paragraph{Ultra-Wideband Ranging.}

UWB is characterized by its large bandwidth and very short waveforms, which is well-suited for ranging applications, such as asset tracking~\cite{ubisense,zhao2021uloc, arun2022realtime}, robotic localization~\cite{ochoa2022landing,lee2022drone, zheng2022uwbgraphopt, cao2020accurate,armani2024accurate} and collaboration~\cite{queralta2022viouwbbased, corrales2008hybridtracking}. 
It is increasingly available in commercial devices such as smartphones, smartwatches, and tags (\eg AirTags)~\cite{samsung, apple, smartposer}.
A challenge in UWB ranging is the noise in non-line of sight (NLOS) conditions. This is especially relevant in human motion tracking, where the human body is an obstacle between two ranging UWB radios. 
Researchers have addressed this by analyzing raw channel impulse response on UWB radios~\cite{barral2019nlos,angarano2021edge, tran2022deepcir} or via sensor fusion with IMUs~\cite{feng2020kalman,ochoa2022landing, mueller2015fusing, hol2009tightlycoupled} or cameras~\cite{queralta2022viouwbbased, Xu2022omniswarm}, and
effectively filtering distance estimates~\cite{armani2024accurate}. 
Building on this approach, \projshort estimate distances in any constellation of trackers, worn by one or multiple people. 

\section{Method}
\label{sec:method}


\subsection{Problem Formulation}

\projshort addresses multi-person pose estimation using sparse inertial sensors and between-sensor distances. 
\projshort is a generic approach that can be applied to any number of users. 
For simplicity and clarity, we consider a scenario with two users in our notation, each equipped with $S=6$ sparse sensors placed on the head, pelvis, wrists and knees.  
Each sensor includes a 6-DoF IMU and a single UWB sensor. For each user $i \in \{1, 2\}$ and at each frame $t$, we obtain 3D orientations \( R^i_t \in \mathbb{R}^{S \times 3} \) and accelerations \( A^i_t \in \mathbb{R}^{S \times 3} \), both represented in a shared world frame. 
Additionally, we denote the pairwise same-person sensor distances as \( D^i_t \in \mathbb{R}^{S \times S} \) for user \( i \), and the between-person distances as \( D^{12}_t \in \mathbb{R}^{S \times S} \) at timestamp \( t \).
Given a sequence of length \( N \) represented by \([A^1, A^2, R^1, R^2, D^1, D^2, D^{12}]\), we predict the SMPL~\cite{loper2015smpl} pose parameters \(\mathbf{\Theta}^i \in \mathbb{R}^{N \times 3J}\) and the translation \(T^i \in \mathbb{R}^{N \times 3}\) for each user \(i\), all in a shared frame of reference. Here,  \( J = 24\)  represents the number of joints.

\subsection{Method Overview}
\Cref{fig:pipeline} shows an overview of our proposed method \projname (\projshort). 
The first step of our pipeline is \textbf{Individual Pose Estimation}, which involves estimating each person’s full-body pose independently using a learning-based pose estimator. 
We design a human pose estimator based on State Space Models~\cite{gu2021efficiently} to produce sequences of full-body SMPL poses $\hat{\Theta}^i$ and root translation trajectories $\hat{T}^i$ using the acceleration, orientation, and same-person sensor distances from six IMUs per person.
To extend from individual pose to multi-person pose, our method introduces two optimization steps:
The \textbf{Initial Position Optimization} refines the initial positions of the two trajectories. 
We define one user's initial position $P_1$ as the world origin $(0, 0, 0)$. 
Our goal is to optimize the second user's position $P_2$, which is relative to $P_1$.
This step aligns trajectories into a shared world frame and provides a stable initialization for the next optimizer.
The \textbf{Trajectory Optimization} further refines the full trajectories by integrating between-human distance $D^{ij}$ constraints through a trajectory optimizer. This step enforces consistency with the relative distances between individuals. 
Note that, directly optimizing the entire trajectory (without the first optimization step) results in unstable convergence to unrealistic paths. 

\begin{figure}[t]
    \centering
    \includegraphics[width=\linewidth]{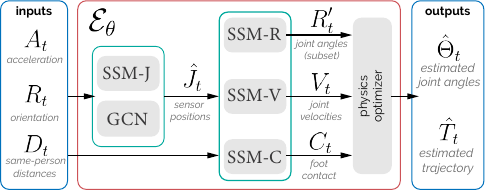}
    \caption{Our SSM-based model takes the global acceleration, rotation, and between-sensor distances to estimate the pose and translation via SMPL.}
    \label{fig:detailed_model}
\end{figure}

\subsection{Individual Pose Estimation}

\projshort starts from individual pose estimation and is performed separately for each user. 
As shown in~\Cref{fig:detailed_model}, the human pose estimator ($\mathcal{E}_\theta$) is state-space model based, which uses acceleration $\mathbf{A}^i$, orientation $\mathbf{R}^i$, and same-person distances $D^i$ to predict SMPL pose parameters (joint orientations) $\mathbf{\hat{\Theta}}^i$~\cite{loper2015smpl} and translations \( \hat{T}^i \) which are relative to their own initial frame:
\begin{equation}
    \mathbf{\hat{\Theta}}^i, \hat{T}^i = \mathcal{E}_\theta(\mathbf{A}^i, \mathbf{R}^i, D^i)
\end{equation}
Since the pose estimator $\mathcal{E}_\theta$ processes each user independently, SMPL poses and translations for each user are estimated separately when multiple users are present.

Inspired by recent advancements in structured state-space sequence models (S4), which have shown remarkable proficiency in sequence modeling~\cite{gu2021efficiently}, we design our pose estimator based on state-space models (SSMs). The S4 model leverages SSMs to efficiently capture complex patterns in sequential data, making it highly effective for modern, large-scale applications. Compared to traditional models like LSTMs~\cite{hochreiter1997long}, S4 offers improved scalability, long-range dependency modeling, and parallelization.
SSMs rely on a classical continuous-time system that maps an input sequence $x(t) \in$ $\mathcal{R}$, through intermediate implicit states $h(t) \in \mathcal{R}^N$ to an output $y(t) \in \mathcal{R}$. The aforementioned process can be formulated as a linear Ordinary Differential Equation (ODE): $h^{\prime}(t) =\mathrm{A} h(t)+\mathrm{B} x(t)$, and $y(t) =\mathrm{C} h(t)$, 
where $\mathrm{A} \in \mathcal{R}^{N \times N}$ denotes the  state transition matrix, while $\mathrm{B} \in \mathcal{R}^{N \times 1}$ and $\mathrm{C} \in \mathcal{R}^{1 \times N}$ represents the projection parameters.
The S4 model discretizes this continuous system, making it suitable for deep learning scenarios. Specifically, it introduces a timescale parameter $\Delta$ and applies fixed discretization rules to transform A and B into discrete parameters $\overline{\mathrm{A}}$ and $\overline{\mathbf{B}}$. Typically, zero-order hold $(\mathrm{ZOH})$ is employed as the discretization rule, defined as follows: $ \overline{\mathrm{A}}=\exp (\Delta \mathrm{A}) $, $\overline{\mathrm{B}}=(\Delta \mathrm{A})^{-1}(\exp (\Delta \mathrm{~A})-\mathrm{I}) \cdot \Delta \mathrm{B}$.
After discretization, the SSM can be computed through linear recurrence, described as

\begin{equation}
\begin{aligned}
& h(t)=\overline{\mathrm{A}} h(t-1)+\overline{\mathrm{B}} x(t), \\
& y(t)=\mathrm{C} h(t)
\end{aligned}
\end{equation}

Our pose estimator consists of four structured SSM~(S4) modules, one graph convolutional network (GCN) module, and a physics optimizer. Since GCNs have been effective in capturing between-sensor distance information~\cite{UIP}, we incorporate a GCN to process orientation and between-sensor distances, predicting sensor positions. Meanwhile, the SSM-J module takes orientation and acceleration as inputs to estimate sensor positions. We learn adaptive weights for SSM-J and GCN to fuse their predictions effectively.
Next, given the predicted sensor positions and the input sensor acceleration, orientation, and same-person sensor distances, SSM-R, SSM-V, and SSM-C predict joint angles, joint velocities, and foot contact states, respectively. Finally, these predicted values are passed to a physics-based optimization module to ensure physical correctness, following~\cite{yi2022physical, UIP}. This produces the final estimated joint angles $\mathbf{\hat{\Theta}}^i$ and translation $\hat{T}^i$.

\subsection{Initial Position Optimization}

\projshort initializes by setting the initial position of the first person, \( T^1_0 \), as the origin in the global coordinate system, while setting the initial position of the second person as \( T^2_0 \). 
Given these two positions, the relative position between them is represented as \(T^{12}_0\).
The objective of this step is to determine the optimal \(T^{12}_0\) by minimizing the differences between two sets of between-person distances: one is derived from \emph{prediction} (denoted as \( PD_t \)) and the other is from \emph{UWB} sensing (denoted as \( D_t \)).
Specifically, the \emph{predicted} distances are computed with the individual pose estimator with forward kinematics.
The \emph{UWB-based distances} are direct measurements. 
Since each user wears 6 sensors, the optimization considers only the $6 \times 6$ between-person sensor pairs, excluding any within-user pairs, to refine the spatial alignment of the two users.

To generate the predicted between-person differences, at each timestamp \( t \), we compute the global positions of all sensors for each person using forward kinematics applied to SMPL parameters \( \mathbf{\hat{\Theta}}^i_t \) and the translation \( T^i_t \), which are relative to each person's initial position \( P_i \). 
The predicted sensor positions, denoted \( SP^i_{t} \in \mathbb{R}^{S \times 3} \), are calculated as follows:

\begin{equation}
SP^i_{t} = P_i + \hat{T}^i_t + fk(\mathbf{\hat{\Theta}}^i_t),
\end{equation}

\noindent
where \( fk(\mathbf{\hat{\Theta}}^i_t) \) is the forward kinematics function that computes the body mesh based on joint rotations and estimates sensor positions from the corresponding mesh vertices.
From the predicted sensor coordinates, we calculate the predicted between-person sensor distances \( PD_t \in \mathbb{R}^{S \times S} \). In detail, each distance is computed as:

\begin{equation}
PD_t(j, k) = \left\| SP^1_t(j) - SP^2_t(k) \right\|_2,
\end{equation}

\noindent
where \( j \) and \( k \) index the sensors on user 1 and user 2, respectively. The predicted distances are compared with the actual distances \( D^{12}_t \in \mathbb{R}^{S \times S} \) obtained from UWB measurements.

To streamline the full process, we define \( FK \) as the composite function that takes SMPL parameters, translations, and initial positions as input, applies forward kinematics via \( fk \), and computes pairwise distances \( PD_t \):
\begin{equation}
    PD_t = FK(\Theta^1, \Theta^2, T^1, T^2, T^{12}_0) 
\end{equation}

\noindent

The goal here is to adjust \( T^{12}_0 \) so that the predicted and actual between-person sensor distances align as closely as possible over the entire trajectory of \( T \) timestamps: 



\small
\begin{equation}
(T^{12}_0) = \underset{\hat{T}^{12}_0} {\arg\min}\sum_{t=1}^{T} \left\| FK(\hat{\Theta}^1_t, \hat{\Theta}^2_t, \hat{T}^1_t, \hat{T}^2_t, \hat{T}^{12}_0) - D^{12}_t \right\|^2_2,
\end{equation}
\normalsize

\noindent

\subsection{Trajectory Optimization}

The initial positions optimization effectively aligns the trajectories of two people into a shared world frame. 
The trajectory optimization step then improves trajectories of both users simultaneously. 


\footnotesize
\begin{align} 
T^1_{t}, T^2_{t} &= \underset{\left(\hat{T}^1_{t}, \hat{T}^2_{t}\right)}{\arg\min} \sum_{t=1}^{T} \left\| FK\left(\hat{\Theta}^1_t, \hat{\Theta}^2_t, \hat{T}^1_t, \hat{T}^2_t, T^{12}_0\right) -  D^{12}_t \right\|_2^2 \\
& \quad + \lambda_1 \sum_{i=1, 2} \left\| \Delta T^i - \Delta \hat{T}^i \right\|_2^2  \quad + \lambda_2 \sum_{i=1, 2} \left\| \Delta^2 T^i - \Delta^2 \hat{T}^i \right\|_2^2
\end{align}
\normalsize

\noindent
In this formulation: The first term (\(7\)) enforces alignment between predicted and observed between-person sensor distances.
The two regularization terms (\(8\)) promote translation smoothness. Let \( \Delta \mathbf{T} \) and \( \Delta^2 \mathbf{T} \) denote the first- and second-order differences (velocity and acceleration) of a person's translation:

\begin{align}
\Delta \mathbf{T} &= \mathbf{T}_{:, 2:T} - \mathbf{T}_{:, 1:T-1} \\
\Delta^2 \mathbf{T} &= \Delta \mathbf{T}_{:, 2:T-1} - \Delta \mathbf{T}_{:, 1:T-2} \\
&= \mathbf{T}_{:, 3:T} - 2 \times \mathbf{T}_{:, 2:T-1} + \mathbf{T}_{:, 1:T-2}
\end{align}

\noindent
One term penalizes the difference between true and predicted velocities(\( \Delta T^i \) vs. \(\Delta \hat{T}^i \)), and the other does the same for accelerations (\( \Delta^2 T^i \) vs. \( \Delta^2 \hat{T}^i\)). Together, they encourage smoother motion by constraining predicted speed and acceleration.

\subsection{Implementation Details} 
We integrate four S4-based neural networks, namely SSM-J, SSM-R, SSM-V, and SSM-C (see Figure~\ref{fig:detailed_model}), each designed to process different output features. These models share a consistent architecture, beginning with a linear encoder that maps the input features to a hidden dimension of 256. The exception is SSM-C, where the hidden dimension is reduced to 32, due to the smaller output size $(2)$. 
Each network then passes through two residual S4 layers, which incorporate LayerNorm and dropout (0.2) for regularization. Finally, a linear decoder maps the output representations to the respective task-specific output spaces.
We train the model on a single NVIDIA 4090 GPU with a batch size of 256, a sequence length of 200, and a learning rate of $1 \times 10^{-3}$, which decays by a factor of 0.33 every 20~epochs. 
The \emph{initial position optimizer} and \emph{trajectory optimizer} utilize the L-BFGS optimizer~\cite{liu1989limited}, with a maximum of four iterations per optimization step and a history size of 10. We use \textit{strong Wolfe} line search to ensure robust step size selection and stable convergence~\cite{liu1989limited}.

\section{Group Inertial Poser Dataset (\datasetname)}
\label{sec:dataset}

\begin{figure}[t]
    \centering
    \includegraphics[width=\linewidth]{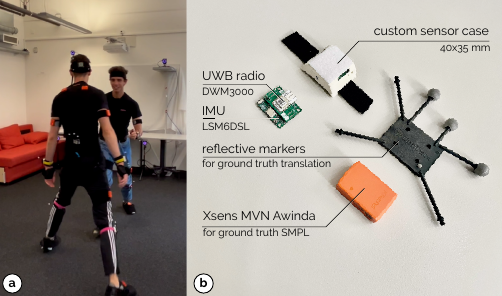}
    \caption{(a) Our dataset includes pairs of participants equipped with (b) various motion capture sensors. These sensors capture acceleration, orientation, and distance data, along with ground-truth SMPL pose parameters and translations for each participant.}
    \label{fig:dataset}
\end{figure}

To evaluate our method on real-world data, we collected a motion capture dataset featuring seven pairs of participants (10 male, 4 female) with heights ranging from 160\, cm to 195\, cm. 
The dataset includes everyday movements such as walking, stretching, and jogging in place, as well as interactive activities like close conversation, sparring, handshakes, and dancing.
Each participant was equipped with an Xsens MVN Awinda suit~\cite{XSens} consisting of 17~IMUs, which provided ground-truth SMPL pose parameters via Xsens' proprietary software. Additionally, they wore six custom wireless sensors placed on the head, pelvis, wrists, and knees. Each custom sensor integrates a UWB radio~(DWM3000) and a 6DoF IMU~(LSM6DSL).
The UWB system, running a ranging protocol and filtering pipeline from existing work~\cite{armani2024accurate}, measured distances at 40 Hz across 12 sensor pairs: (15x2)~same-person sensor distances and 36~between-person sensor distances. 
Meanwhile, each IMU recorded acceleration $A_t$ and orientation $R_t$ at 104 Hz.
We additionally captured ground-truth translation using a 20-camera OptiTrack system~\cite{optitrack}, using the pelvis sensor as the reference point.
All sensors were calibrated according to their respective documentation~\cite{XSens, armani2024accurate}. Each recording session began and ended with a T-pose and a jump to synchronize data across sources (Xsens, OptiTrack, and custom sensors). Figure~\ref{fig:dataset} illustrates our setup.
In our \datasetname dataset, the average UWB RMSE is 5 cm for same-person measurements, increasing to 15 cm for between-person measurements due to greater occlusions.

\begin{table*}[t]
\small
    \setlength{\tabcolsep}{4pt}
    \centering
    \caption{Results for training on AMASS and evaluation on the InterHuman dataset. 
    When directly comparing with PIP and UIP, we assume the initial position is known for all methods (upper table), as PIP and UIP cannot predict the initial relative translations. We also report results when \projshort is not initialized with ground truth but using our \textit{initial position optimizer}, which only affects the translation errors and shows comparable performance to the ground truth initialization. 
    The best results are in \textbf{bold}.}
    \label{tab:inter_human_no_noise}
    \begin{tabular}{lccccccccc}
    \toprule
    Method & SIP Err  & Angle Err  & Joint Err & Vertex Err &   Dist Err-4s&   Dist Err-8s&Dist Err-12s&Dist Err-16s &Dist Err-20s\\
     & (°)  &  (°)   &  (cm)  & (cm) &   (cm)&   (cm)&(cm)&(cm) &(cm)\\
    \midrule
    PIP + D & 20.95 & 14.89 & 7.98 & 9.40 &   48.82&   58.25&59.71&63.39&63.72\\

    UIP & 19.25 & 12.89 & 6.60 & 7.31 &   49.61&   62.51&59.10&83.16&81.44\\

    \projshort (ours)  & \textbf{18.30} & \textbf{9.94}& \textbf{5.74} & \textbf{6.53} &   \textbf{3.08}&   \textbf{3.73}&\textbf{4.40}&\textbf{1.36}&\textbf{1.91}\\
    
    \midrule
    \projshort (ours init.opt.) & \textbf{18.30} & \textbf{9.94}& \textbf{5.74} & \textbf{6.53} &   3.19&  4.00 & 4.70 &4.90&\textbf{1.91}\\
    
    \bottomrule
    \end{tabular}
\end{table*}





\begin{table*}[t]
\small
    \setlength{\tabcolsep}{4pt}
    \centering
    \caption{
    Results for training on AMASS data and evaluation on the real-world \datasetname dataset. When directly comparing with PIP and UIP, we assume the initial position is known for all methods, as PIP and UIP cannot predict the initial translation.\vspace{-2mm}}
    \label{tab:inter_uwb}
    \begin{tabular}{lccccccccc}
    \toprule
    Method & SIP Err  & Angle Err  & Joint Err & Vertex Err & Dist Err-4s& Dist Err-8s&  Dist Err-12s&Dist Err-16s &Dist Err-20s\\
     & (°)  &  (°)   &  (cm)  & (cm) & (cm)& (cm)&  (cm)&(cm)  &(cm)\\
    \midrule
    PIP + D & 30.55 & 27.40 & 11.43 & 12.37 & 33.22& 37.83&  46.65&54.56&73.70\\

    UIP & 30.18 & 26.16 & 10.88 & 11.50 & 31.11& 37.49&  44.24&55.73&74.93\\

    \projshort (ours) & 27.77 & 23.34& 9.45 & 10.21 & \textbf{23.06}&23.79&  \textbf{21.86}& 19.82&20.69\\

    \midrule
    \projshort (ours init.opt.) & 27.77 & 23.34& 9.45& 10.21& 23.57& 24.40&  22.79& \textbf{19.34}&20.71\\
    \projshort (finetuned) & \textbf{18.04} & \textbf{17.57} & \textbf{8.70} & \textbf{9.60} & 23.36& \textbf{23.27}& 22.07& 19.56&\textbf{19.59}\\
    \bottomrule
    \end{tabular}
\end{table*}


\section{Experiments}
\label{sec:experiments}

To assess the advantages of \emph{\projshort} for multi-human inertial pose estimation, we conduct experiments on both synthetic and real-world data. We compare our method against previous inertial sensing-based approaches, namely PIP~\cite{yi2022physical} and UIP~\cite{UIP}.
There are fundamental differences between \projshort and these prior methods. Notably, neither PIP nor UIP was designed to estimate inter-human translation. To address this limitation, we initialize their translation at frame zero using the ground truth.
Additionally, PIP was originally designed to work solely with IMU data and does not leverage between-sensor distances. To ensure a fair comparison, we provide PIP with additional sensor distance measurements following the same approach as UIP~\cite{UIP}.

\noindent\paragraph{Datasets.} 
Our evaluation aims to show the benefits of \emph{\projshort} in both synthetic and real-world scenarios. 
All methods are trained from scratch to estimate individual poses using   synthesized AMASS data~\cite{AMASS:ICCV:2019}.  
Specifically, we generate synthetic data by computing IMU measurements (acceleration and rotation) and between-sensor distances on virtual sensors, following prior work~\cite{huang2018deep, yi2022physical, UIP}.
We use two datasets for evaluation: InterHuman~\cite{liang2024intergen} and our real-world \datasetname dataset, both of which contain two-person interactions.
For InterHuman, we synthesize IMU and UWB data using the same procedure as for AMASS.




\noindent\paragraph{Metrics.} 
Following previous work on human pose estimation from inertial sensors~\cite{yi2022physical, yi2021transpose, UIP}, we evaluate the predicted poses and trajectory of SMPL using the following metrics:
\textit{SIP Error (°)} evaluates the global joint angle error of arms (shoulder angle) and legs (hip angle). 
\textit{Angle Error (°)} evaluates all global joint angle errors. 
\noindent\textit{Joint Error ($cm$)} evaluates the root aligned mean per joint position error (MPJPE). 
\noindent\textit{Trans Error @\{3m, 6m\} ($cm$)} The global root translation error is computed over all movement pairs that span a distance of $x$ meters. This metric quantifies the deviation from the true trajectory over fixed-distance intervals. Indicates the diversion from the true trajectory over time.
We introduce \textit{Dist Err-\{4s, 8s, 12s, 16s, 20s\}($cm$)}, which captures the distance error between two individuals over the corresponding time. It quantifies the multi-human relative translation error, providing more meaningful insights compared to global translation errors for interaction scenarios.


\subsection{Evaluation Results}
\begin{figure}[t]
    \centering
    \includegraphics[width=\columnwidth]
    {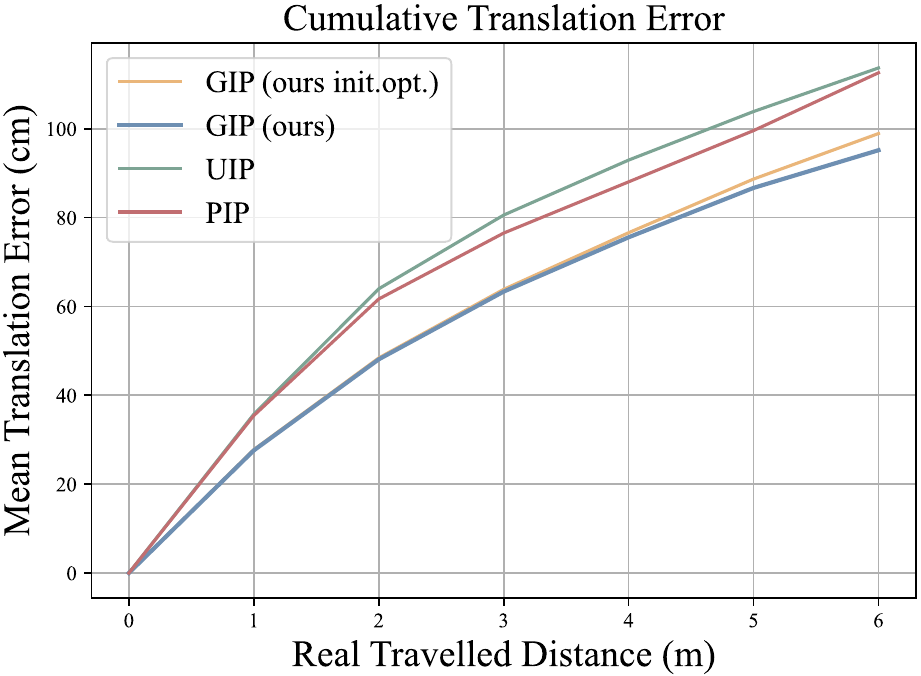}%
    \vspace{-2mm}%
    \caption{Comparison of translation error on InterHuman.
}%
    \label{fig:trans_error}%
    \vspace{-4mm}%
\end{figure}

\vspace{-5mm}
\noindent\paragraph{Quantitative Results on the InterHuman Dataset.}
As shown in \Cref{tab:inter_human_no_noise}, \emph{\projshort} outperforms the baselines across all pose and translation metrics. Our model, incorporating trajectory optimization, produces significantly more accurate and consistent relative translations over time. Additionally, our SSM-based pose estimation method reduces full-body joint angle error by 22\% compared to UIP and 33\% compared to PIP. 
Notably, when initialized using the \textit{initial position optimizer}, \projshort closely matches the performance of the ground-truth-initialized version, further validating the effectiveness of our approach. 
\Cref{fig:trans_error} shows that our method achieves the lowest cumulative translation error. 

\noindent\paragraph{Quantitative Results on the \datasetname Dataset.}
Compared to finite difference-based simulations of accelerations or distances between virtual sensors, \datasetname's recordings of \emph{actual} UWB distances and IMU values are noisier as it was recorded from real-world behavior.
This in turn makes accurate predictions more challenging. 
\Cref{tab:inter_uwb} highlights \emph{\projshort's} benefits over PIP and UIP when dealing with real world noisy data.
We improve all pose and translation metrics by a substantial margin, specifically, we reduce the distance error by 72\% at 20s.  
Again, the initial pose optimizer proves its benefit and yields comparable results as the ground truth initialized experiment.
We also show that when we use the \datasetname data to finetune the model, the model performance on angular prediction could be significantly improved.

\begin{figure*}[t]
    \centering%
    \vspace{-1mm}%
\includegraphics[width=\linewidth]{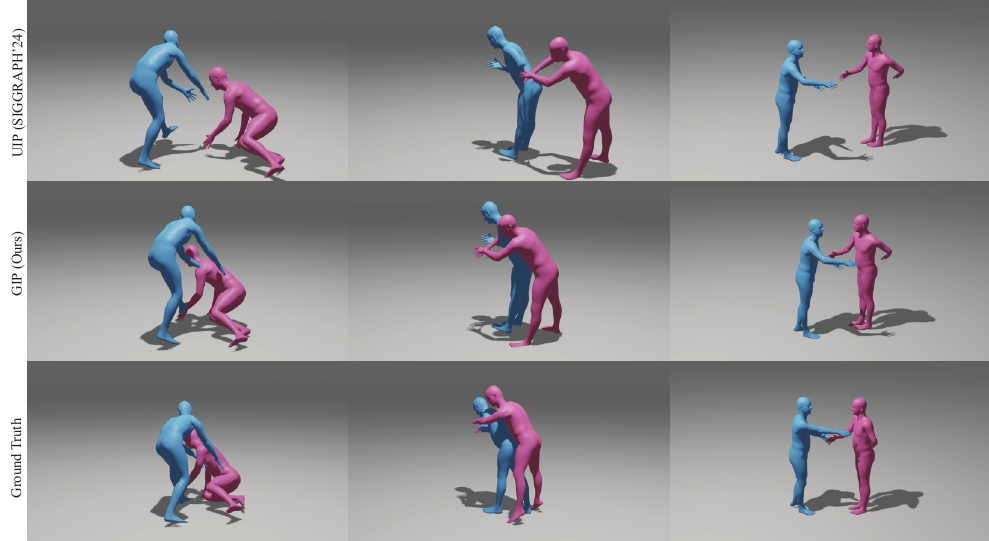}
    \caption{
    Visual comparison of \emph{\projshort} and \emph{UIP}. \emph{\projshort} effectively corrects trajectory errors and preserves inter-personal interaction dynamics.
    }%
    \vspace{-2mm}%
\label{fig:qualitative_results}%
\end{figure*}

\noindent\paragraph{General Qualitative Results.} 
\Cref{fig:qualitative_results} presents a visual comparison of our proposed method, \emph{\projshort}, and the state-of-the-art method, \emph{UIP}. It is clear that \emph{UIP} struggles to estimate relative translations accurately, making it difficult to capture inter-personal interactions. In contrast, \emph{\projshort} effectively improves both the relative translations between individuals and the global translations.

\vspace{-2mm}
\noindent\paragraph{Multi-User Scenarios.}

%
We conducted an experiment on a synthetic four-person EgoHumans~\cite{khirodkar2023ego} dataset. 
We keep the test subject fixed and add more people to the optimization process. For example, with three people—p1, p2, and p3—we jointly optimize the group (p1, p2, p3). When a fourth person, p4, is added, we perform two separate optimizations: first on (p1, p2, p3), then on (p1, p2, p4). 
Adding more people improves translation estimation for the same person thanks to the additional spatial constraints (Table~\ref{tab:people_trans_error}).


\begin{table}[h]
    \small
    \caption{
    Translation error decreases with more people involved.
    }%
    \vspace{-2mm}%
    \centering
    \setlength{\tabcolsep}{10.0pt}
    \begin{tabular}{lcccc}
    \toprule
         $N_\text{people}$ &  1 &  2 & 3 & 4 \\
    \midrule
         Trans Error (m)&  2.34 &  1.52&  1.20 & 1.15\\
    \bottomrule
    \end{tabular}
    \label{tab:third_person}
    \vspace{-4mm}
    \label{tab:people_trans_error}
\end{table}

\begin{table}[t]
    \setlength{\tabcolsep}{3pt}
    \centering
    \caption{
    Ablation study on the InterHuman~\cite{liang2024intergen} testset. 
    }%
    \vspace{-2mm}%
    \label{tab:ablat}
    \resizebox{\columnwidth}{!}{
    \begin{tabular}{lcccccc}
    \toprule
    Method & SIP Err  & Angle Err  & Trans-3m   & Trans-6m & RMSE & MAE\\
     & (°)  &  (°)   & (cm)  & (cm) & (cm) &(cm)\\
    \midrule
    w/o init. opt. & 18.30 &9.94 & 99.50 & 116.01 &76.50 & 69.16\\
    w/o traj. opt. & 18.30 &9.94 & 76.07 & 115.32 & 40.00& 36.21 \\
    w/LSTM & 19.24 & 12.88 & 66.60 & 103.74 & 42.19 & 38.53\\
    \textbf{Ours} & \textbf{18.30} & \textbf{9.94}&\textbf{ 63.85 } &  \textbf{98.93}\textbf{ } & \textbf{34.86} & \textbf{31.11}\\

    \bottomrule
    \end{tabular}
    }
\end{table}

\section{Ablation Studies}
\label{sec:ablation}
\Cref{tab:ablat} shows the impact of \emph{\projshort}'s components on pose and translation accuracy. All evaluations assume that the initial position is unknown. 
The result highlights the importance of the \emph{initial position optimization}, as removing this step significantly degrades translation accuracy (1\textsuperscript{st} row). 
The trajectory optimizer further improves the translations in a shared frame, improving translation accuracy (2\textsuperscript{nd} row), reducing the translation error @6m by up to 16\,cm. 
To assess the effectiveness of SSM, we substitute SSMs with LSTMs (4\textsuperscript{th} row) while keeping the rest of the pipeline intact (optimizations included). We demonstrate that LSTMs perform worse in both angle errors and translations.

\section{Limitations and Discussion}
\label{sec:limitations}


Our method has demonstrated significant improvements over prior approaches for inertial-based motion capture across multiple datasets. Nevertheless, several limitations remain. First, as noted in prior work~\cite{UIP}, the UWB noise remains a significant challenge, particularly in multi-person interactions where signal obstruction is common.
Nonetheless, our results suggest that future advancements in UWB precision could further enhance motion accuracy. 
Second, our approach estimates body pose while assuming a mean body shape, without accounting for inter-individual shape variation. Future research could extend this work by incorporating body shape estimation from sparse observations, as explored in~\cite{jiang2024egoposer}.
Third, the use of optimization-based inference introduces computational overhead. Although our method converges in fewer than 10 iterations and processes a 30-second motion sequence in 2.04 seconds, this remains a limiting factor for resource-constrained applications. 
Finally, our method does not explicitly mitigate foot sliding, which can partially arise from our trajectory optimization.





\section{Conclusion}
\label{sec:conclusion}

Accurate multi-person tracking using sparse sensing is an essential step toward generalized motion tracking and capturing meaningful inter-personal interactions. 
For this purpose, \emph{\projname} overcomes the drift and lack of positional references in previous inertial methods and demonstrates multi-person motion tracking by augmenting inertial measurements with between-sensor distances.
We leverage these novel constraints to mitigate drift and improve relative translation accuracy.
Beyond improved single-person pose estimation with a novel SSM model, \projshort robustly tracks two individuals using sparse IMU+UWB sensors and accurately estimates relative trajectories.
Unlike previous methods, \projshort allows people to start from arbitrary positions and automatically determines their initial relative positions via a two-step optimization.
\projshort offers quantitative and qualitative advantages---it improves accuracy but also preserves crucial inter-personal interaction dynamics. 
Additionally, we introduce \textit{\datasetname}, the first IMU+UWB dataset designed for multi-person tracking using sparse inertial sensors.
Evaluated with InterHuman and \datasetname, our approach consistently outperforms existing methods in accuracy and robustness across synthetic and real-world data.
Collectively, our work highlights the potential of IMU+UWB fusion for multi-person motion tracking, opening new opportunities for real-world applications.




\section*{Acknowledgments}
We thank all participants of our data capture.
We thank Adnan Harun Dogan for his help with preparing the dataset.
Yi-Chi Liao was supported by the ETH Zurich Postdoctoral Fellowship Program.
Ying Xue was in part supported by the Swiss National Science Foundation (Grant No.~10004941).

{
    \small
    \balance
    \bibliographystyle{ieeenat_fullname}
    \bibliography{main}
}

\end{document}